\documentclass[conference]{IEEEtran}
\IEEEoverridecommandlockouts
\usepackage{cite}
\usepackage{amsmath,amssymb,amsfonts}
\usepackage{stfloats}   
\usepackage{algorithmic}
\usepackage{mdwtab}
\usepackage{multirow}
\usepackage{tabularx}
\usepackage{graphicx}
\usepackage{textcomp}
\usepackage{subfig}
\usepackage{xcolor}

\def\BibTeX{{\rm B\kern-.05em{\sc i\kern-.025em b}\kern-.08em
    T\kern-.1667em\lower.7ex\hbox{E}\kern-.125emX}}
\begin{document}

\title{A Capsule Network for Traffic Speed Prediction \\ 
		in Complex Road Networks}

\author{\IEEEauthorblockN{Youngjoo Kim, Peng Wang, Yifei Zhu, and Lyudmila Mihaylova}
\IEEEauthorblockA{\textit{Department of Automatic Control and Systems Engineering} \\
\textit{The University of Sheffield}\\
Sheffield, United Kingdom \\
\{youngjoo.kim, peng.wang, yzhu42, l.s.mihaylova\}@sheffield.ac.uk }}

\maketitle

\begin{abstract}
This paper proposes a deep learning approach for traffic flow prediction in complex road networks. Traffic flow data from induction loop sensors are essentially a time series, which is also spatially related to traffic in different road segments. The spatio-temporal traffic data can be converted into an image where the traffic data are expressed in a 3D space with respect to space and time axes. Although convolutional neural networks (CNNs) have been showing surprising performance in understanding images, they have a major drawback. In the max pooling operation, CNNs are losing important information by locally taking the highest activation values. The inter-relationship in traffic data measured by sparsely located sensors in different time intervals should not be neglected in order to obtain accurate predictions. Thus, we propose a neural network with capsules that replaces max pooling by dynamic routing. This is the first approach that employs the capsule network on a time series forecasting problem, to our best knowledge. Moreover, an experiment on real traffic speed data measured in the Santander city of Spain demonstrates the proposed method outperforms the state-of-the-art method based on a CNN by 13.1\% in terms of root mean squared error.
\end{abstract}


\section{Introduction}
Traffic prediction is one of the central tasks for building intelligent transportation management systems in metropolitan areas. Traffic congestion causes delays and costs millions to the economy worldwide, which is worse in urban centres. Predicting the traffic flow will provide the stakeholders with tools for modelling and decision support.

Early approaches to traffic flow prediction are statistical techniques including support vector machines (SVM) \cite{Wu2004} and the autoregressive integrated moving average (ARIMA) model \cite{Williams2003}. These statistical approaches have been demonstrated to be effective as regression techniques for time series data. However, they do not address the spatio-temporal relationship of transportation networks and cannot be applied to a large-scale road network. Recently, machine learning technologies \cite{Lv2015, Zhang2017, Ma2015, Wu2018, Ma2017} have been actively applied given that traffic prediction is essentially to make estimations of future state based on big data. The spatio-temporal features of the traffic has been of great interest of researchers. Understanding the spatial evolution of traffic for the entire road network rather than for a small part of the network is necessary at both stages of off-line planning and on-line traffic management. Convolutional neural networks (CNNs) have been successful in dealing with spatial features of road networks \cite{Lv2015, Zhang2017}. Besides, recurrent neural networks (RNNs) with long short-term memories (LSTM) \cite{Zhang2017, Ma2015} and gated recurrent unit (GRU) \cite{Wu2018} have been incorporated, considering the traffic flow prediction as a time series forecasting.

A novel approach has been proposed in \cite{Ma2017} that converts the traffic speed into images where the traffic speed data of each road segment at each time step is expressed in the third dimension. A CNN is used to capture spatio-temporal features in the images. This method has been demonstrated to outperform other state-of-the-art methods. This approach differs from others in that other approaches simply treat the time dimension of the traffic flow as a channel of image data and therefore the temporal features of traffic flow are ignored \cite{Wu2018}. However, this work was demonstrated on rectangular sub-networks of a metropolitan road network, which have a relatively simple topology.

\begin{figure}[t!]
	\centering
	\includegraphics[width=0.95\linewidth]{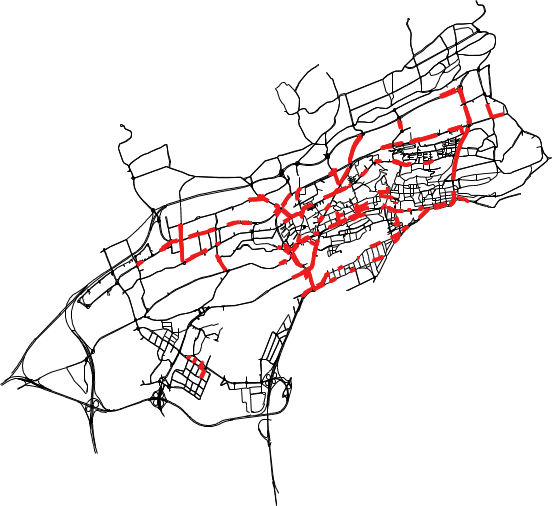}
	\caption{Road network of central Santander city. Red lines denote road segments where the speed sensors are located.}
	\label{fig01}
\end{figure}

The goal of this study is to devise a traffic speed prediction method for complex road networks. We are dealing with traffic data gathered in a metropolitan area of Santander city of Spain, whose road network is depicted in Fig. 1. The red lines denote road segments where induction loop detectors are installed to measure vehicle speeds. The speed sensors are sparsely located in a complex network. Adjacency in the spatio-temporal image does not necessarily mean adjacency in the road network. In this case, some spatial features would be disregarded by a CNN because it uses the max pooling operation to construct higher order features locally. Therefore, we utilize a capsule network (CapsNet) \cite{Hinton2018, Sabour2017} that replaces the pooling operation with dynamic routing, which enables to take into account important spatial hierarchies between simple and complex objects. We propose a CapsNet architecture designed to be suitable for traffic speed prediction and demonstrate its effectiveness by comparing it with the CNN-based method in \cite{Ma2017}. To our best knowledge, this is the first application of the CapsNet to a time series forecasting problem.

The rest of this paper is organized as follows. It starts with addressing the method of converting traffic data into images in Section II. After that, an existing approach to traffic speed prediction based on a CNN is introduced. We then present the proposed architecture of the CapsNet designed for traffic speed prediction. The methods and results of the performance evaluation with a real dataset are given in Section III. Finally, Section IV presents a summary and conclusion.

\section{Traffic Speed Prediction}

\subsection{Traffic Speed Data as an Image}

\begin{figure}[t]
	\centering
	\includegraphics[width=0.95\linewidth]{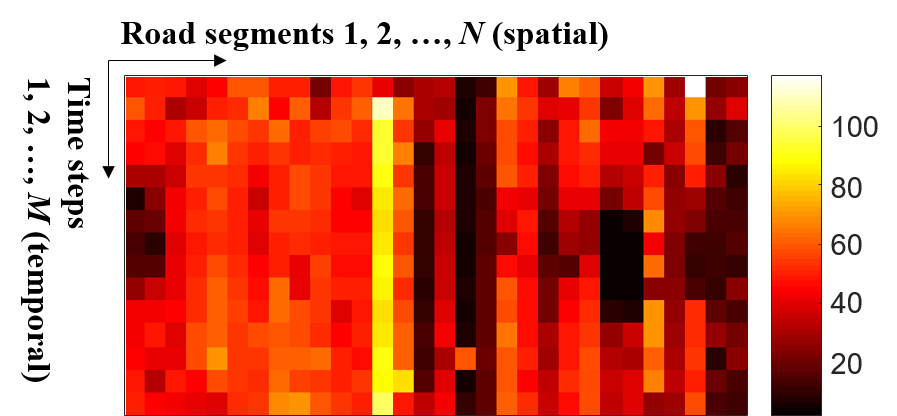}
	\caption{Spatio-temporal image representation of traffic speed data (unit: km/h).}
	\label{fig02}
\end{figure}

\begin{figure*}[t]
	\centering
	\includegraphics[width=0.95\linewidth]{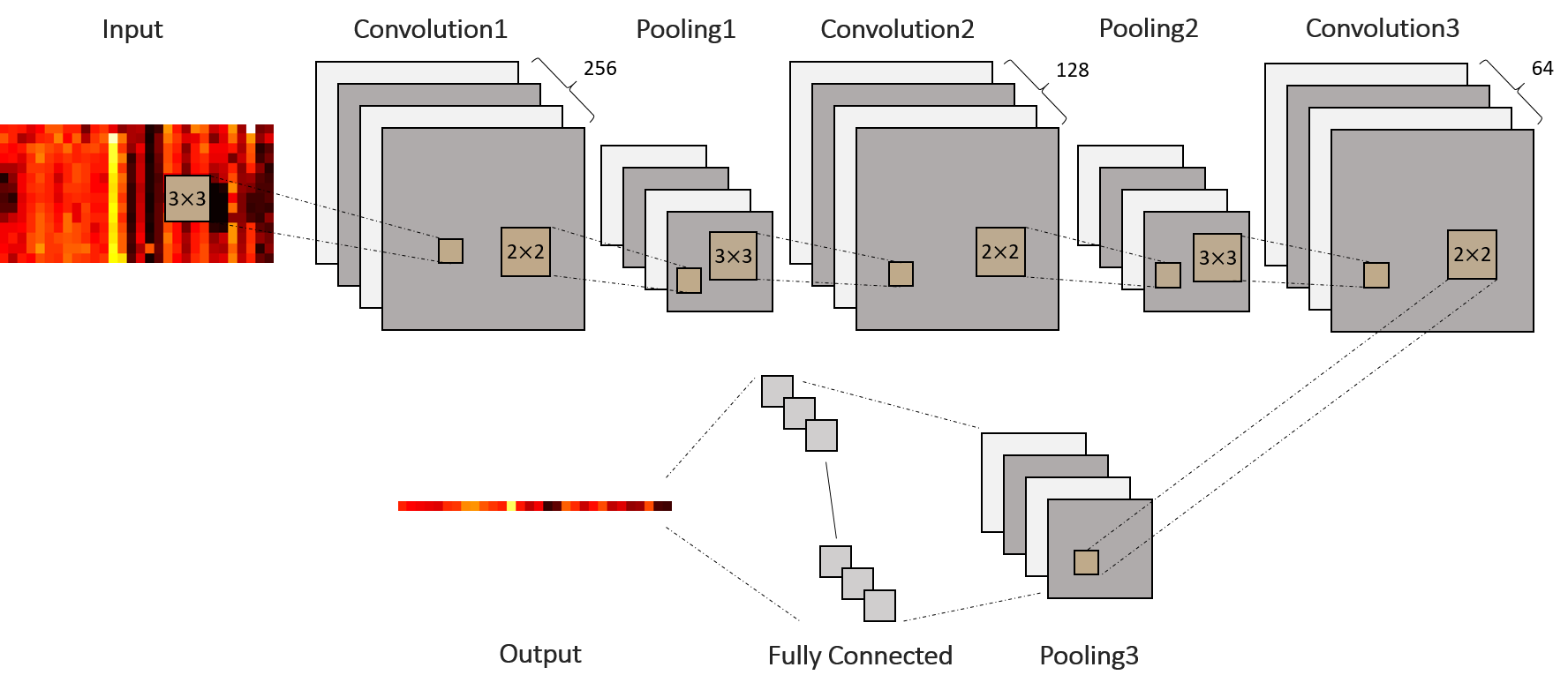}
	\caption{Architecture of CNN for traffic speed prediction.}
	\label{fig03}
\end{figure*}

\begin{figure*}[b]
	\centering
	\includegraphics[width=0.95\linewidth]{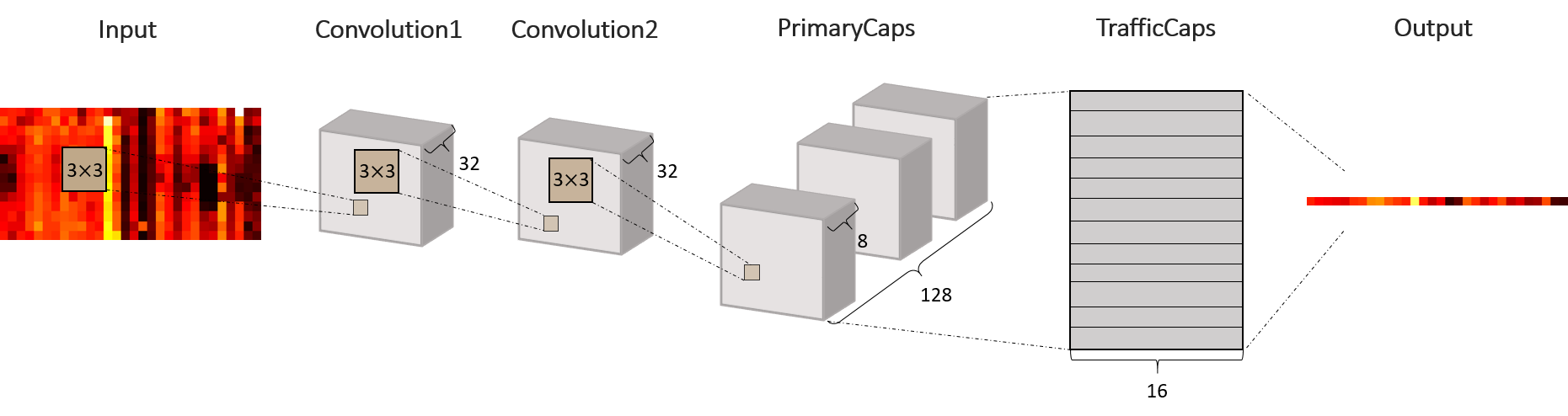}
	\caption{Architecture of CapsNet for traffic speed prediction.}
	\label{fig04}
\end{figure*}

Each induction loop sensor records time history of traffic speed on different road segments. In order to consider the spatio-temporal relationship, the traffic data are converted to an image with two axes representing time and space. As a result, we have an image as an $M{\times}N$ matrix where $M$ and $N$ denote the number of time steps and the number of sensors, respectively. The matrix is then represented as:
\begin{equation}
X = \begin{bmatrix}
	x_{11} & \cdots & x_{1N} \\
	\vdots & \ddots & \vdots \\
	x_{M1} & \cdots & x_{MN}
\end{bmatrix}
\end{equation}
where $x_{mn} (m=1,...,M, \; n = 1,...,N)$ denotes the traffic speed at $m$-th time step in $n$-th road segment. For example, Fig. 2 depicts the spatio-temporal image representation of traffic speed data. 

Suppose we have traffic speed data from $N$ sensors and we are going to predict the traffic speed in $L$ time steps ahead based on data from previous $M$ time steps. Given the overall time history of traffic speed, a strip of $M{\times}N$ matrix becomes an input and the data in the next time steps act as labels in training neural networks. The output can be an array with a size of $LN$, which can be obtained by reshaping a strip of $L\times N$ matrix to an array as:
\begin{equation}
Y = \begin{bmatrix}
y_1 & \cdots & y_N & y_{N+1} & \cdots & y_{LN}
\end{bmatrix}
\end{equation}

\subsection{CNN for Traffic Speed Prediction}

The CNN has been demonstrated to be significantly effective in understanding images by using max pooling and successive convolutional layers that reduce the spatial size of the data flowing through the network. These procedures increase the field of view of high-level layers and allow them to capture high-order features of the input image.

\begin{table}[t!]
	\renewcommand{\arraystretch}{1.5}
	\caption{Layer parameters of CNN.}
	\label{table1}
	\centering
	\begin{tabular}{| c | c | c |}
		\hline
		\textbf{Layer} & \textbf{Parameter} & \textbf{Activation} \\
		\hline
		Convolution1 & (256, 3, 3) & ReLu \\
		\hline
		Pooling1 & (2, 2) & - \\
		\hline
		Convolution2 & (128, 3, 3) & ReLu \\
		\hline
		Pooling2 & (2, 2) & - \\
		\hline
		Convolution3 & (64, 3, 3) & ReLu \\
		\hline
		Pooling3 & (2, 2) & - \\
		\hline
		Flattening & - & - \\
		\hline
		Fully-connected & - & - \\
		\hline
	\end{tabular}
	\renewcommand{\arraystretch}{1}
\end{table}

\begin{table}[b!]
	\renewcommand{\arraystretch}{1.5}
	\caption{Layer parameters of CapsNet.}
	\label{table2}
	\centering
	\begin{tabular}{| c | c | c |}
		\hline
		\textbf{Layer} & \textbf{Parameter} & \textbf{Activation} \\
		\hline
		Convolution1 & (32, 3, 3) & ReLu \\
		\hline
		Convolution2 & (32, 3, 3) & ReLu \\
		\hline
		\multirow{2}{*}{PrimaryCaps} & (128, 3, 3) & ReLu \\
		& Capsule size 8 & - \\
		\hline
		TrafficCaps & Capsule size 16 & - \\
		\hline
	\end{tabular}
	\renewcommand{\arraystretch}{1}
\end{table}

We use the CNN architecture proposed in \cite{Ma2017} as the baseline. This consists of three pairs of a convolutional layer and a pooling layer followed by a flattening operation and a fully-connected layer. Fig. 3 depicts the architecture of the CNN for traffic speed prediction. The three convolutional layers have 256, 128, and 64 channels, respectively, of size 3$\times$3. Each convolution layer involves a rectified linear unit (ReLu) activation function to give nonlinearity to the network.  

Pooling layers have filters of size 2$\times$2 applied with a stride of 2. This downsamples every depth slice in the input by 2 and reduces the redundancy of representation by removing 75\% of the activations. The output of each max pooling filter is determined by taking the maximum over 4 numbers in a 2$\times$2 region. The output of the last pooling layer is transformed to a vector by the flattening operation and this contains the final and the highest-level features of the input traffic history. Lastly, the flattened output goes through a fully-connected layer to provide the prediction. The output of the fully-connected layer now has the same dimension as the label vector in (2). The parameters of the CNN is presented in Table I.

\subsection{Proposed CapsNet Architecture}

CNNs have worked surprisingly well in various applications. Nonetheless, max pooling in CNNs is losing valuable information by just picking the neuron with the highest activation. CapsNet has been proposed in \cite{Hinton2018, Sabour2017} to address the drawback of CNNs.

A capsule is a group of neurons that encodes the probability of detection of a feature as the length of their output vector. Each layer in a CapsNet contains many capsules that represent different properties of the same object. One of the main characteristics of capsules is that capsules have vector forms and their activations provide vector outputs whereas artificial neurons go through scalar operations. More importantly, the CapsNet is trained by an algorithm called dynamic routing proposed in \cite{Sabour2017}. The dynamic routing is executed between two successive capsule layers to update weights that determine how the low-level capsules send their input to the high-level capsules that agree with the input. In other words, the weights are determined based on the dot product of the low-level capsule and the high-level capsule where the dot product captures the similarity of two vectors. Each weighted sum of the low-level capsules is then passed through the squash function that forces the length to be no more than 1 while preserving the direction of the vector. Unlike CNNs, the CapsNet does not throw away information that is most likely relevant to the task at hand, like relative relationships between spatio-temporal traffic features. 

In the proposed architecture, as depicted in Fig. 4,  the first two convolutional layers convert the spatio-temporal traffic image to the activities of local feature detectors used as inputs to the third layer. The third layer, called PrimaryCaps, is another convolutional layer that has 128 channels with a 3$\times$3 kernel. All the convolution operations are performed with a stride of 1 with zero padding, involving a ReLu nonlinearity. Each capsule in the PrimaryCaps layer is an 8-dimensional vector and capsules in a cuboid are sharing their weights with each other. The final layer, called TrafficCaps, has a 16-dimensional capsule per road segment. The dynamic routing is performed between PrimaryCaps and TrafficCaps with 3 iterations. The dynamic routing algorithm captures the relationship between all the capsules in the PrimaryCaps layer and each capsule representing each road segment. In this way, any distant local feature can contribute to characterizing the capsules in the TrafficCaps layer. Here we consider the length of each 16-dimensional capsule vector in the TrafficCaps layer as the traffic speed on the corresponding road segment. The parameters of the proposed CapsNet are given in Table II. 

\section{Performance Validation with Real Data}

\begin{figure}[t!]
	\centering
	\includegraphics[width=0.95\linewidth]{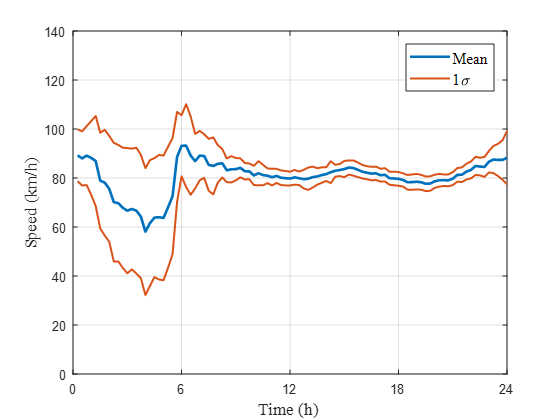}
	\caption{Mean and 1-sigma variation of all 1-year data on a road segment.}
	\label{fig05}
\end{figure}

We use traffic speed data measured every 15 minutes on road segments in the central Santander city for a year of 2016. The dataset is from the case studies of the SETA EU project \cite{SETA2016}. Excluding days when the sensors did not work, the spatio-temporal traffic dataset is a matrix with a size of 33054$\times N$ where $N$ denotes the number of road segments. Each sparsely missing measurement is masked with an average of measurements taken at the same time in the other days. We use traffic data from January to September as a training set and the remaining data from October to December as an evaluation set. As an example, the average speed and a 1-year variation on a road segment are presented in Fig. 5. Note that each road segment would have different statistics and no topological information of the road network is given. Understanding and predicting the spatio-temporal relationship of traffic between different road segments in different time slots are the duty of the neural networks. The CNN and CapsNet described in Section II. B) and II. C), respectively, performed the following four prediction tasks:

\begin{itemize}
\item Task 1: 15-min prediction with 150-min traffic history on 20 road segments ($L=1, M=10, N=20$)
\item Task 2: 30-min prediction with 150-min traffic history on 20 road segments ($L=2, M=10, N=20$)
\item Task 3: 15-min prediction with 210-min traffic history on 50 road segments ($L=1, M=14, N=50$)
\item Task 4: 30-min prediction with 210-min traffic history on 50 road segments ($L=2, M=14, N=50$)
\end{itemize}

\begin{figure}[t!]
	\centering
	\subfloat[Case 1]{%
		\includegraphics[clip,width=0.95\linewidth]{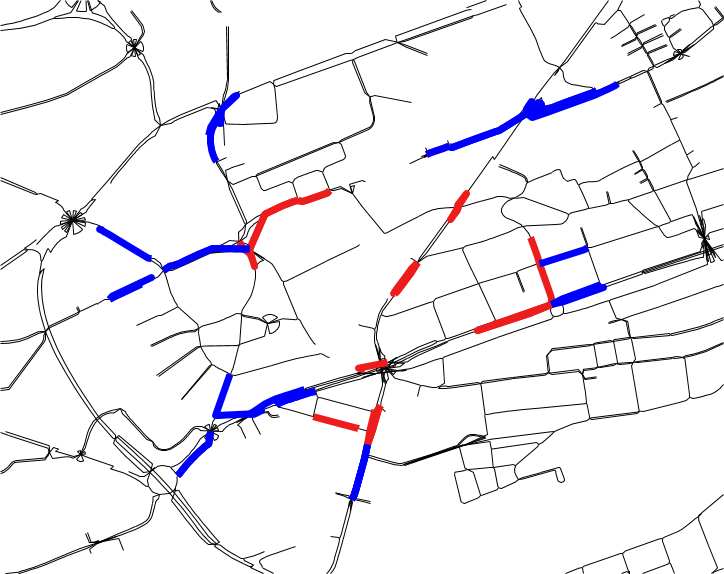}%
	}
	
	\subfloat[Case 2]{%
		\includegraphics[clip,width=0.95\linewidth]{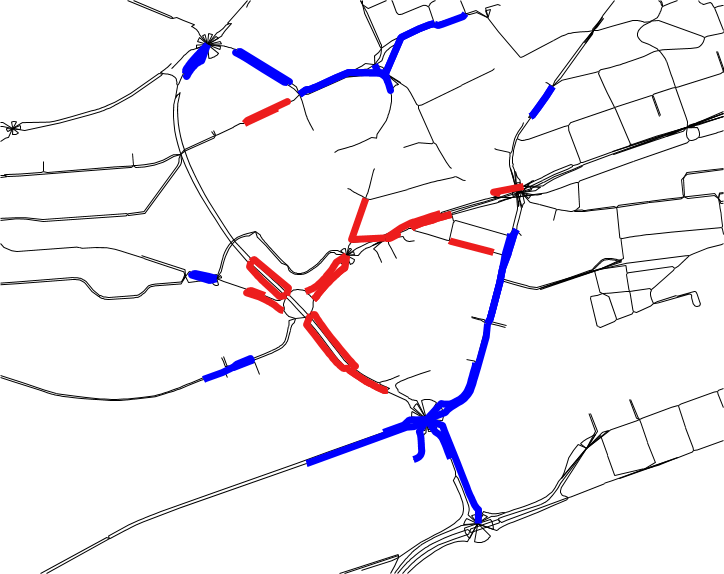}%
	}
	
	\caption{Road segments used in the experiments. The first 20 segments are marked in red and the adjacent 30 segments are marked in blue.}
	\label{fig06}
\end{figure}

The traffic prediction tasks are performed in two sets of road segments as depicted in Fig. 6. 20 road segments used in Task 1 and Task 2 are marked in red and the other 30 road segments, used in Task 3 and Task 4 together with the red segments, are marked in blue. Note that traffic data in adjacent road segments are not always located close to each other in the spatio-temporal image. We attempt to verify the methods on larger spatio-temporal images in Task 3 and Task 4 where the neural networks are required to capture the spatio-temporal features scattered in a larger region.

In our Tensorflow implementation, each network employs mean squared error (MSE) as a loss function and we use the Adam optimizer \cite{Kingma2014} with the exponentially decaying learning rate to minimize the sum of the MSE. We scale the traffic speed data into the range [0,1] before feeding in the neural networks. 

The prediction result can be compared with the true values in the form of images. Fig. 7 depicts the image representation of the true traffic speed and predictions by the CapsNet and the CNN. Traffic speed data at 3 different time periods are drawn where the images in the same column represent the traffic data at the same time period. It is observed that the deep learning methods provide similar results as if a smoothing filter is applied to the true traffic images.

\begin{figure}[t!]
	\centering
	\subfloat[True]{%
		\includegraphics[clip,width=0.9\linewidth]{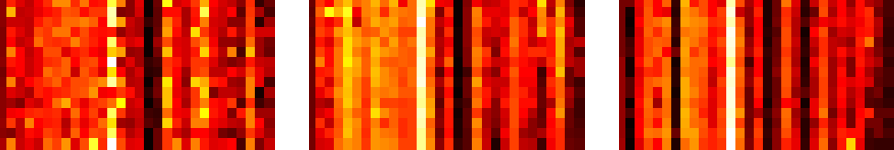}%
	}
	
	\subfloat[CapsNet]{%
		\includegraphics[clip,width=0.9\linewidth]{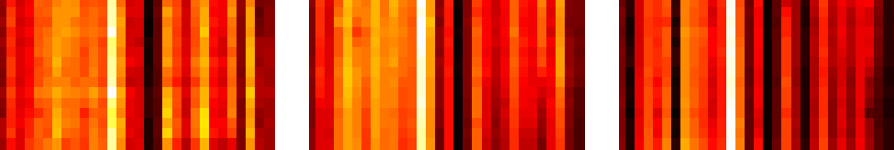}%
	}
	
	\subfloat[CNN]{%
		\includegraphics[clip,width=0.9\linewidth]{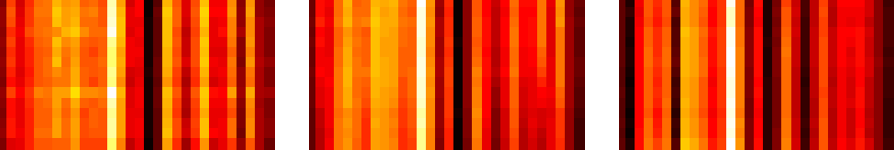}%
	}
	
	\caption{Prediction result for traffic speed compared in the form of images.}
	\label{fig07}
\end{figure}

The images shown in Fig. 7 are just snapshots of the result. Since we have a lot of data in the evaluation set, statistical performance metrics are required to assess the overall performance of the networks. Mean relative error (MRE) is one of the most common metric to quantify accuracy of different prediction models in general. However, the error of a smaller value of speed might result in larger MRE and vice versa. Thus, we further employ mean absolute error (MAE) and root mean squared error (RMSE) as more intuitive metrics for assessing the speed prediction performance. The three performance metrics are defined as:
\begin{equation}
MRE = \frac {\sum_{i=1}^{I} { |y_i - \hat{y}_i|/y_i } } {I} 
\end{equation}
\begin{equation}
MAE = \frac {\sum_{i=1}^{I} { |y_i - \hat{y}_i| } } {I} 
\end{equation}
\begin{equation}
RMSE = \sqrt { \frac{\sum_{i=1}^{I}{ (y_i - \hat{y}_i)^2 }}  {I} }
\end{equation}
where $\hat{y}_i$ and $y_i$ denote the $i$-th speed prediction and its true value, respectively. Here, $I$ represents the number of the speed data in the evaluation set.

\begin{table}[t!]
	\captionsetup{justification=centering}
	\caption{Prediction performance \\ (unit: \% for MRE, km/h for MAE and RMSE).}
	\label{table3}
	\centering
	\renewcommand{\arraystretch}{1.5}
	\subfloat[Case 1]{%
		\begin{tabular}{|c|c|c|c|c|c|c|}
			\hline
			\multirow{2}{*}{} & \multicolumn{3}{c|}{\textbf{CNN}}                                      & \multicolumn{3}{c|}{\textbf{CapsNet}}                                  \\ \cline{2-7} 
			& \textit{\textbf{MRE}} & \textit{\textbf{MAE}} & \textit{\textbf{RMSE}} & \textit{\textbf{MRE}} & \textit{\textbf{MAE}} & \textit{\textbf{RMSE}} \\ \hline
			\textbf{Task 1}   & 5.668                 & 6.102                 & 10.30                  & 0.444                 & 5.675                 & 8.853                  \\ \hline
			\textbf{Task 2}   & 0.649                 & 6.204                 & 10.47                  & 0.289                 & 5.791                 & 9.179                  \\ \hline
			\textbf{Task 3}   & 18.14                 & 6.323                 & 10.68                  & 5.146                 & 5.790                 & 9.257                  \\ \hline
			\textbf{Task 4}   & 4.661                 & 6.583                 & 10.85                  & 0.876                 & 5.898                 & 9.472                  \\ \hline
		\end{tabular}
	}

	\subfloat[Case 2]{%
		\begin{tabular}{|c|c|c|c|c|c|c|}
			\hline
			\multirow{2}{*}{} & \multicolumn{3}{c|}{\textbf{CNN}}                                      & \multicolumn{3}{c|}{\textbf{CapsNet}}                                  \\ \cline{2-7} 
			& \textit{\textbf{MRE}} & \textit{\textbf{MAE}} & \textit{\textbf{RMSE}} & \textit{\textbf{MRE}} & \textit{\textbf{MAE}} & \textit{\textbf{RMSE}} \\ \hline
			\textbf{Task 1}   & 37.35                 & 6.519                 & 10.98                  & 1.555                 & 6.109                 & 9.362                  \\ \hline
			\textbf{Task 2}   & 21.41                 & 6.667                 & 11.19                 & 10.97                 & 6.240                 & 9.718                  \\ \hline
			\textbf{Task 3}   & 19.76                 & 6.915                 & 11.29                  & 9.746                 & 6.113                 & 9.674                  \\ \hline
			\textbf{Task 4}   & 2.333                 & 6.957                 & 11.38                  & 4.146                 & 6.243                 & 9.913                  \\ \hline
		\end{tabular}
	}

	\renewcommand{\arraystretch}{1}
\end{table}

The performance of the CNN and CapsNet has been assessed with their best settings. Both of the networks show their best performance with the common starting learning rate of 0.0005 and the exponential decay rate of 0.9999. The resultant performance of the neural networks on the four tasks with two datasets is presented in Table III. The MRE does not seem to provide consistent results. On the other hand, the MAE and RMSE increase as the input and output sizes increase from Task 1 to Task 4. The CapsNet shows better (smaller) MAE and RMSE than the CNN in all the tasks in both cases. The performance difference is larger in Task 3 and Task 4 where the size of the input image is larger. The CapsNet provided 6.58\% smaller MAE in Task 1 and Task 2 and 10.2\% smaller MAE in Task 3 and Task 4. We conclude the CapsNet is better at capturing the relationship between distant spatio-temporal features as expected. In average, the CapsNet provides 8.24\% and 13.1\% improvement in MAE and RMSE, respectively, compared with the CNN.

A drawback of the CapsNet is that it takes a longer time to train the network. In our experiment of Task 1, the CapsNet is about 30 times slower than the CNN. The computation time difference becomes severe for tasks with larger output sizes. The number of trainable parameters in the CapsNet varies from $8.24\times10^6$ (Task 1) to $143\times10^6$ (Task 4) whereas that in the CNN varies from $0.374\times10^6$ (Task 1) to $0.410\times10^6$ (Task 4). Given increased input and output sizes, the routing algorithm requires a significant increase in the number of trainable parameters because it deals with a full-scale image features by testing all the combinations between multidimensional vectors, called capsules. On the other hand, the number of trainable parameters shows a mere increase in the CNN, which is contributed by the pooling operation.

\section{Conclusion}
This paper presents a capsule net framework that captures the spatio-temporal features of traffic speed and provides short-term traffic speed predictions. The vehicular traffic speed measured by magnetic loop detectors is represented as images that are fed into the developed capsule network. Traffic speed predictions by the proposed CapsNet architecture are compared with those by a CNN-based method. Experiments performed on 1-year data measured on road segments in Santander city demonstrate the proposed CapsNet provides more accurate speed predictions than the CNN. The performance difference is larger in experiments with a larger dataset. This result implies the CapsNet is better at learning spatio-temporal features in the test data, with 13.1\% improvement in RMSE with respect to the CNN.

\section*{Acknowledgment}
The authors appreciate the support of the SETA project funded by the European Union’s Horizon 2020 research and innovation program under grant agreement no. 688082.

\bibliography{Library_SSDF}
\bibliographystyle{IEEEtran}

\end{document}